\documentclass[conference]{IEEEtran}
\IEEEoverridecommandlockouts
\usepackage{cite}
\usepackage{amsmath,amssymb,amsfonts}
\usepackage{algorithmic}
\usepackage{graphicx}
\usepackage{textcomp}
\usepackage{subcaption}
\usepackage{subfloat}
\usepackage{threeparttable}
\usepackage{xcolor}
\usepackage{multirow}
\usepackage{booktabs}
\usepackage{makecell}
\usepackage{amsthm}
\usepackage{algorithmic}
\usepackage{hhline}
\usepackage[ruled,linesnumbered]{algorithm2e}
\def\BibTeX{{\rm B\kern-.05em{\sc i\kern-.025em b}\kern-.08em
    T\kern-.1667em\lower.7ex\hbox{E}\kern-.125emX}}
\begin{document}

\title{DanZero+: Dominating the GuanDan Game through Reinforcement Learning
}
\author{Youpeng Zhao,
        Yudong Lu,
		Jian Zhao,
		Wengang Zhou,
		Houqiang Li
}
\maketitle

\begin{abstract}
The utilization of artificial intelligence (AI) in card games has been a well-explored subject within AI research for an extensive period. 
Recent advancements have propelled AI programs to showcase expertise in intricate card games such as Mahjong, DouDizhu, and Texas Hold'em.
In this work, we aim to develop an AI program for an exceptionally complex and popular card game called GuanDan.
This game involves four players engaging in both competitive and cooperative play throughout a long process to upgrade their level, posing great challenges for AI due to its expansive state and action space, long episode length, and complex rules.
Employing reinforcement learning techniques, specifically Deep Monte Carlo (DMC), and a distributed training framework, we first put forward an AI program named DanZero for this game.
Evaluation against baseline AI programs based on heuristic rules highlights the outstanding performance of our bot.
Besides, in order to further enhance the AI's capabilities, we apply policy-based reinforcement learning algorithm to GuanDan.
To address the challenges arising from the huge action space, which will significantly impact the performance of policy-based algorithms, we adopt the pre-trained model to facilitate the training process and the achieved AI program manages to achieve a superior performance.

\end{abstract}

\section{Introduction}
Adopting games as benchmark environments for evaluating the efficacy of artificial intelligence (AI) has sparked notable interest within the machine learning community, particularly in the realm of reinforcement learning \cite{pinto2018hierarchical, silva2018strategy, perez2019general}.
Recent progress in reinforcement learning has resulted in substantial advancements across diverse games, encompassing strategic board games like Go \cite{silver2016mastering, silver2017mastering} and chess \cite{silver2018general}, card games such as Texas Hold'em \cite{heinrich2016deep, moravvcik2017deepstack, brown2018superhuman} and tile-based games like Mahjong \cite{li2020suphx}.
Additionally, progress has been witnessed in the domain of video games, exemplified by achievements in StarCraft \cite{vinyals2019grandmaster} and DOTA \cite{berner2019dota}.
Nevertheless, challenges still exist in games characterized by imperfect information and extensive state and action spaces, presenting formidable hurdles for reinforcement learning methodologies.

\begin{figure*}[t]
	\centering
     \includegraphics[width=0.95\textwidth]{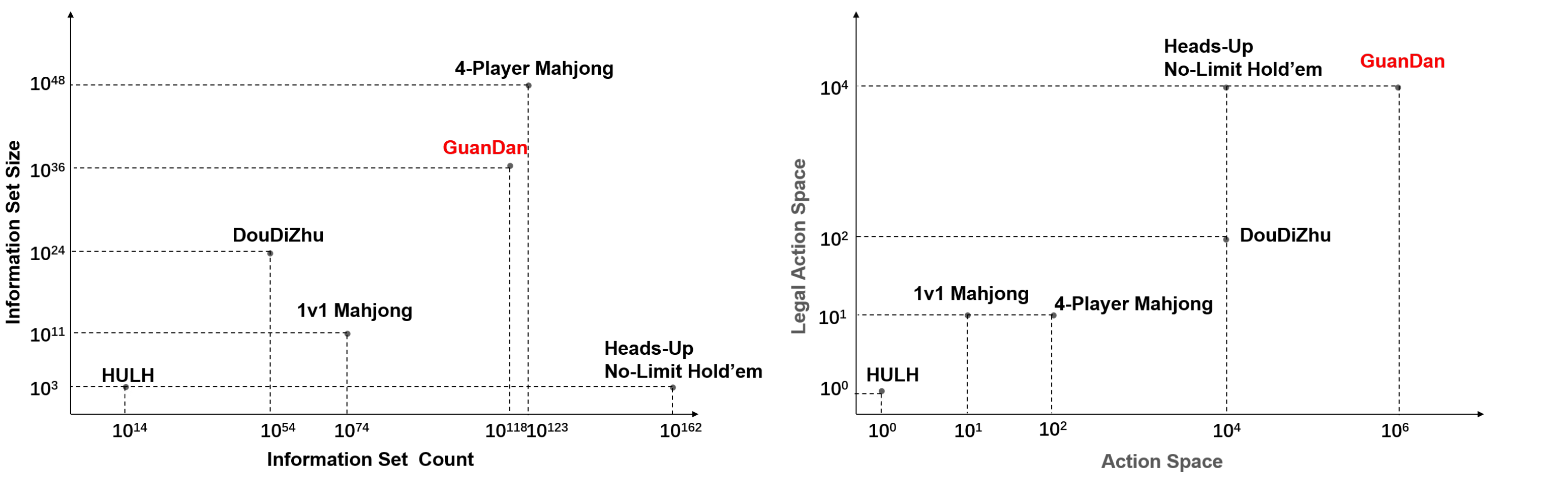}
	\caption{Game complexity of some imperfect information games, including Heads-Up Limit Texas Hold'em (HULH), Heads-Up No-Limit Texas Hold'em, 1v1 Mahjong, 4-Player Mahjong, DouDizhu and GuanDan. }
	\label{fig1}
\end{figure*}

This work is dedicated to developing an artificial intelligence (AI) program tailored for GuanDan, a card game of substantial popularity in China, with over 20 million active participants, yet receiving limited scholarly attention. 
This game involves intricate dynamics of cooperation and competition within a partially observable environment, with four players organized into two groups engaging in gameplay with two decks of cards. 
The inherent complexity of GuanDan emanates from its extensive state and action space, evidenced by a notably augmented information set size and count in comparison to other games, including 4-player Mahjong. 
Additionally, the game also exhibits a considerably larger action and legal action space, illustrated in Figure~\ref{fig1}. 
Despite notable advancements in AI applied to chess and card games\cite{li2020suphx, you2020combinatorial, jiang2019deltadou, zha2021douzero, zhao2022douzero+}, the development of an AI program tailored for GuanDan remains an intricate and unresolved challenge.

In general, some characteristics of GuanDan make this game challenging for developing an AI program:
\begin{itemize}
\item \textbf{The state space and action space of GuanDan is large:} 
    In a GuanDan game, a distinguishing feature is the utilization of two decks of pokers, setting it apart from many games that have been studied which only use one deck of cards. 
    What's more, the significance of card suits is nonnegligible in GuanDan due to their impact on the combination of cards.
    In contrast, such a factor is usually overlooked when dealing with some card games such as DouDizhu.
    In addition, there exist ``wild card'' and ``level card'' in GuanDan, rendering the gameplay considerably more intricate when compared to conventional card games.
    \item \textbf{The length of one episode of GuanDan game is long:} 
    In the context of GuanDan, a notable element is the incorporation of a ``leveling up'' concept.
    Specifically, the game ends only when a team of players has upgraded over level A so that one episode of the game contains several rounds.
    Within each episode, each agent in GuanDan has to execute over 100 decision-making while the number of decisions that agent in other card games needs to make per episode seldom exceeds 20.
    The substantial increase in decision complexity reflects the challenges posed by the intricate dynamics of GuanDan gameplay to some degree.
    \item \textbf{The number of players in the GuanDan game may change throughout the game progression:} Initially, this game involves four players, with two forming a collaborative team pitted against the remaining pair.
    One round of the GuanDan game terminates only when both players of any team have emptied their respective hand cards, making this game very complex.
    For example, if one player has emptied his hand cards, the ensuing game transforms into an imbalanced situation, \emph{i.e.} two players cooperate against the rest single player.
    \item \textbf{The number of legal actions under a state is uncertain:}
    The intricate nature of GuanDan game also arises from the presence of unique card combinations and the ``wild card'' concept.
    Consequently, there can be more than 5000 legal actions at the initial state of a game, but this number may undergo a rapid reduction to fewer than 50 as cards have been played from the hand. This feature poses a considerable challenge in effectively modeling actions within the game.
\end{itemize}

The prevailing approaches in existing AI programs for GuanDan mainly rely on heuristic rules as well as meticulous techniques tailed for different scenarios.
For instance, strategic considerations may prompt a player to prioritize cards that are small or unable to form special card types when leading the first trick.
When a player plays cards passively to counter the moves of other players, using cards of the same card type is prior to playing bomb cards. 
Notably, an attempt \cite{shen2020imperfect} tried to develop an AI program for GuanDan utilizing the Upper Confidence Bound Apply to Tree (UCT) algorithm but it performs only marginally better than random agents, where agents choose actions randomly from legal action set.
Moreover, considering the extensive state and action space of this game, classical value-based reinforcement learning algorithms such as Deep Q-Learning (DQN) \cite{mnih2015human-level} will encounter issues such as overestimating \cite{zahavy2018learn}. 
Similarly, policy gradient methods like A3C \cite{mnih2016asynchronous} exhibit suboptimal performance when confronted with large action spaces, mainly due to their inability to effectively leverage action features.
The intricate nature of GuanDan presents formidable challenges in developing AI systems directly employing conventional reinforcement learning algorithms.

In this study, we propose to develop an AI system for GuanDan employing reinforcement learning methodologies.
First, considering the issues discussed above, we incorporate Monte Carlo methods enhanced by deep neural networks (DMC), which can utilize the action features and approximate true values without bias \cite{sutton2018reinforcement}.
Additionally, we utilize feature encoding techniques to process the state and action features and implement a distributed self-play reinforcement learning framework to facilitate training.
The integration of these techniques helps our AI program outperform other existing algorithms.
Furthermore, we enhance the performance of this AI system by integrating the proximal policy optimization (PPO) algorithm \cite{schulman2017proximal}.
Utilizing the pre-trained model allows us to surmount the challenges associated with huge action space, leading to a stronger AI program for GuanDan: DanZero+.

The preliminary version of this work is documented in \cite{lu2022danzero} and accepted by IEEE Conference on Games 2023. 
We have made significant improvements and extensions to our preliminary work.
The major technical improvement is applying the PPO algorithm to the GuanDan game.
Specifically, we adopt the model trained using the DMC method to provide some candidate actions for the PPO to make decisions, thus successfully addressing the challenges brought by the extensive action space of this game.
In this way, we manage to achieve an AI program with enhanced performance compared to our preliminary GuanDan AI.
To be noted, our idea can also be applied to other policy-based reinforcement learning algorithms.
Besides, we make revisions to some statements in our preliminary work to ensure more accurate and fitting descriptions and demonstrate the outstanding performance of our AI system in this game, affirming the effectiveness of our methods.

\section{Related Work}
This section provides an overview of imperfect information games, emphasizing the application of reinforcement learning methodologies in game AI in such contexts.

Imperfect information games, characterized by hidden information and stochasticity, present a more realistic portrayal of real-world scenarios. Consequently, addressing these games raises more intricate and significant research queries compared to perfect-information games. 
Classical imperfect information games, such as poker, often employ Counterfactual Regret Minimization (CFR) \cite{neller2013introduction} and its variants, necessitating a game model to traverse the expansive game tree during computation. 
As for addressing large-scale imperfect information games, techniques like learning state or action space abstractions to reduce the game's complexity have been widely applied \cite{bowling2015heads, moravvcik2017deepstack, brown2019deep}. 
However, the unique complexity of GuanDan, involving both cooperation and competition with changing player dynamics, presents substantial challenges to traditional poker game algorithms. 
In spite of efforts to utilize deep neural networks for generalization across states in CFR methods, which helps them obviate the need for explicit abstractions \cite{Li2020Double, steinberger2019single}, these approaches still encounter difficulties when handling games with extensive state and action spaces.

While CFR methods heavily rely on game-tree traversals, the paradigm of reinforcement learning introduces a distinctive approach by enabling models to acquire skills through interactions with the environment. 
This adaptability makes reinforcement learning particularly well-suited for addressing the challenges posed by large-scale games.
Recent achievements in reinforcement learning have sparked a growing trend in applying this technique to the domain of imperfect information games.
The successful stories of reinforcement learning tackling famous large-scale games such as DOTA \cite{berner2019dota}, StarCraft \cite{vinyals2019grandmaster} and Honor of King \cite{ye2020mastering} underscore its potential.
It is noteworthy that these notable achievements leverage the Proximal Policy Optimization (PPO) algorithm.
Card games, including Mahjong \cite{li2020suphx}, Texas Hold'em \cite{heinrich2016deep, brown2018superhuman}, and DouDizhu \cite{zha2021douzero, zhao2022douzero+}, have also witnessed the triumphant integration of reinforcement learning techniques.
Besides, reinforcement learning can be easily integrated with complementary techniques such as search algorithms \cite{brown2020combining} and opponent modeling \cite{he2016opponent, knegt2018opponent}, consistently achieving impressive performance. 
Considering these advantages and successful cases, we choose reinforcement learning to develop an AI program for the unresolved GuanDan game.

As for the GuanDan game, current approaches predominantly rely on heuristic rules or CFR methods to build the Guandan AI. 
These schemes mainly incorporate human priority knowledge and stipulate that intelligent bodies make reasonable decisions in various cases. 
At present, the most capable Guandan AI mainly involves hand cards classification, manual value assignment to legal actions based on circumstances, and action selection through value comparisons.
Most of the current solutions do not consider using reinforcement learning algorithms to improve Guandn AI skills.
In contrast, we resort to reinforcement learning techniques to handle this problem.
To be specific, we adopt Deep Monte Carlo (DMC), using neural networks for function approximation in the Monte Carlo method, as the primary framework to train our AI system.
Additionally, drawing inspiration from the success of the Proximal Policy Optimization (PPO) algorithm in large-scale Real-Time Strategy (RTS) and Multiplayer Online Battle Arena (MOBA) games, we also leverage this method to further augment the performance of our AI system.

\section{Preliminary}
In this section, we give a brief introduction to the basic rules of GuanDan, including the card-playing phase and tribute phase, to enhance the comprehension of our research context. 
A more detailed introduction is presented in the appendix.
For a comprehensive set of rules, interested readers may refer to the detailed documentation available on Wiki~\footnote{https://en.wikipedia.org/wiki/Guandan\#Playing}.
Following this, we offer a brief review of the concept and formulation of Proximal Policy Optimization.

\begin{figure}[t]
	\centering
	\includegraphics[width=0.9\columnwidth]{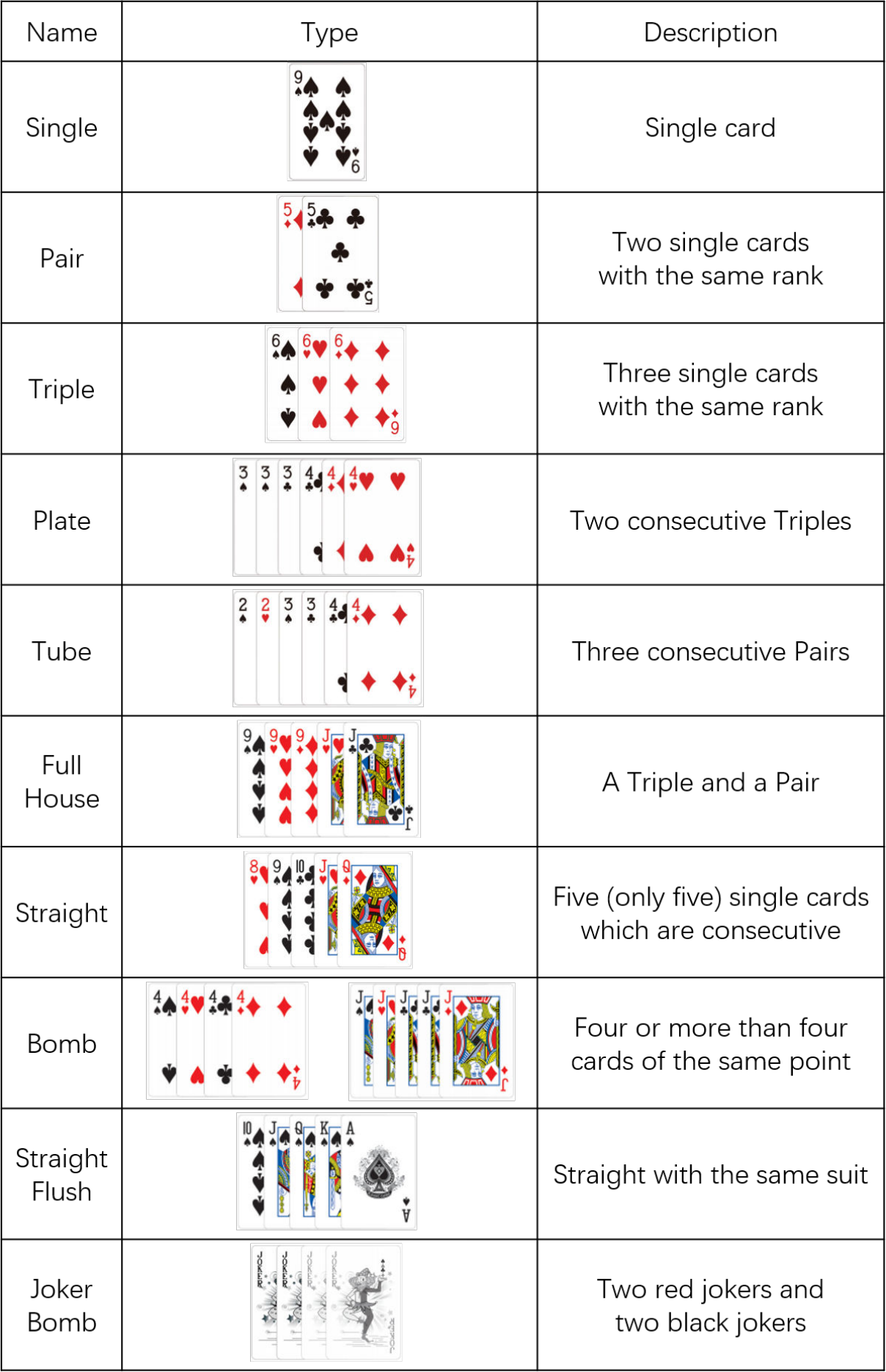}
	\caption{A list of all card types in GuanDan.}
	\label{fig2}
\end{figure}

\subsection{Basic Card Play Knowledge}
\label{chap:rule}
The card types of the GuanDan game are listed in Figure~\ref{fig2}. 
There are four suits in the cards used in the GuanDan game, including Hearts (H), Spades (S), Diamonds (D), and Clubs (C). 
The cardinal rank of single cards is, organized in descending order, Red Joker (RJ), Black Joker (BJ), Level cards, A, K, Q, J, 10, 9, 8, 7, 6, 5, 4, 3, 2. 
Notably, when forming Tubes, Plates, Straights, or Flushes, Aces can be seen as 1, positioned just below 2. 
The regulations stipulate that for Solos, Pairs, Triples, Tubes, Plates, Full Houses, and Straights, cards are confined to covering counterparts of the same type. 
Full Houses are ranked by the points of the Triple part. 
Bombs can cover the aforementioned combinations and Bombs with more cards can cover Bombs with fewer cards. 
In cases where two Bombs possess an identical number of cards, their ranking is further determined by their respective points.  
Flush Straights can cover Bombs with fewer than six cards; the relationship between this kind of card type is determined by the points. 
Finally, Joker Bombs can beat any other card type in the game.

\subsection{Rules in Guandan}
\label{rules}
In a GuanDan game, there are two decks of standard pokers used, including Jokers, and four players are each dealt with a hand comprising 27 cards. 
This gameplay includes two opposing camps, with players seated opposite each other being affiliated with the same faction. 
Further complexity is introduced through the notions of ``leveling up'', ``level cards'', and ``wild cards''.
Cards of the same rank as the ongoing round's level are called ``level cards'' and they hold a rank just below Jokers when being played individually.
Moreover, these cards maintain their natural rank order when contributing to the composition of other combinations.
Particularly noteworthy are Heart cards of the level rank, which are defined as ``wild cards.''
They can substitute for any required cards in combination formation, excluding Jokers.
The overarching objective in GuanDan is for a camp to be the first to achieve a level surpassing `A', thereby clinching victory.
To this end, one GuanDan game usually encompasses multiple rounds, each marked by strategic interplay and nuanced level dynamics.

\begin{figure}[t]
	\centering
	\includegraphics[width=0.92\columnwidth]{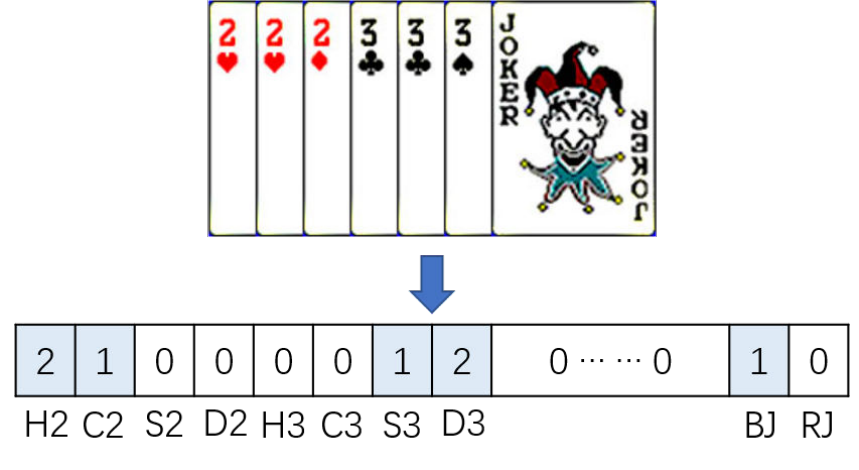}
	\caption{An example to show the encoding of one hand. `H' is the abbreviation of Heart. `C' is the abbreviation of Club. `S' is the abbreviation of Spade and `D' is the abbreviation of Diamond. `BJ' is the abbreviation of Black Joker and `RJ' is the abbreviation of Red Joker.
	0, 1, 2 represents the number of such cards.}
	\label{fig3}
\end{figure}

In a GuanDan game, the player achieving the first card depletion is called ``Banker'', while other players are designated the ``Follower'', the ``Third'', and the ``Dweller'' based on the sequential order of card depletion.
Only the Banker's team can promote the level and the promoted number ranges from three to one, determined by the specific order of card depletion for the Banker's partner.
If the winning team empties the hand cards in the first and second positions, they will be able to promote three levels and their opponents are called the Double-Dweller.

From the second round onward, prior to the beginning of the initial trick, the Dweller of the preceding round is obliged to pay a Tribute to the Banker by relinquishing his biggest single card, excluding the wild card.
In reciprocation, the Banker needs to return a single card with a point not surpassing 10 to ensure that each player has the same number of cards.
Then the Dweller can lead the first trick.
More special cases of the tribute phase are described in the appendix.

\begin{figure*}[t]
	\centering
	\includegraphics[width=0.95\textwidth]{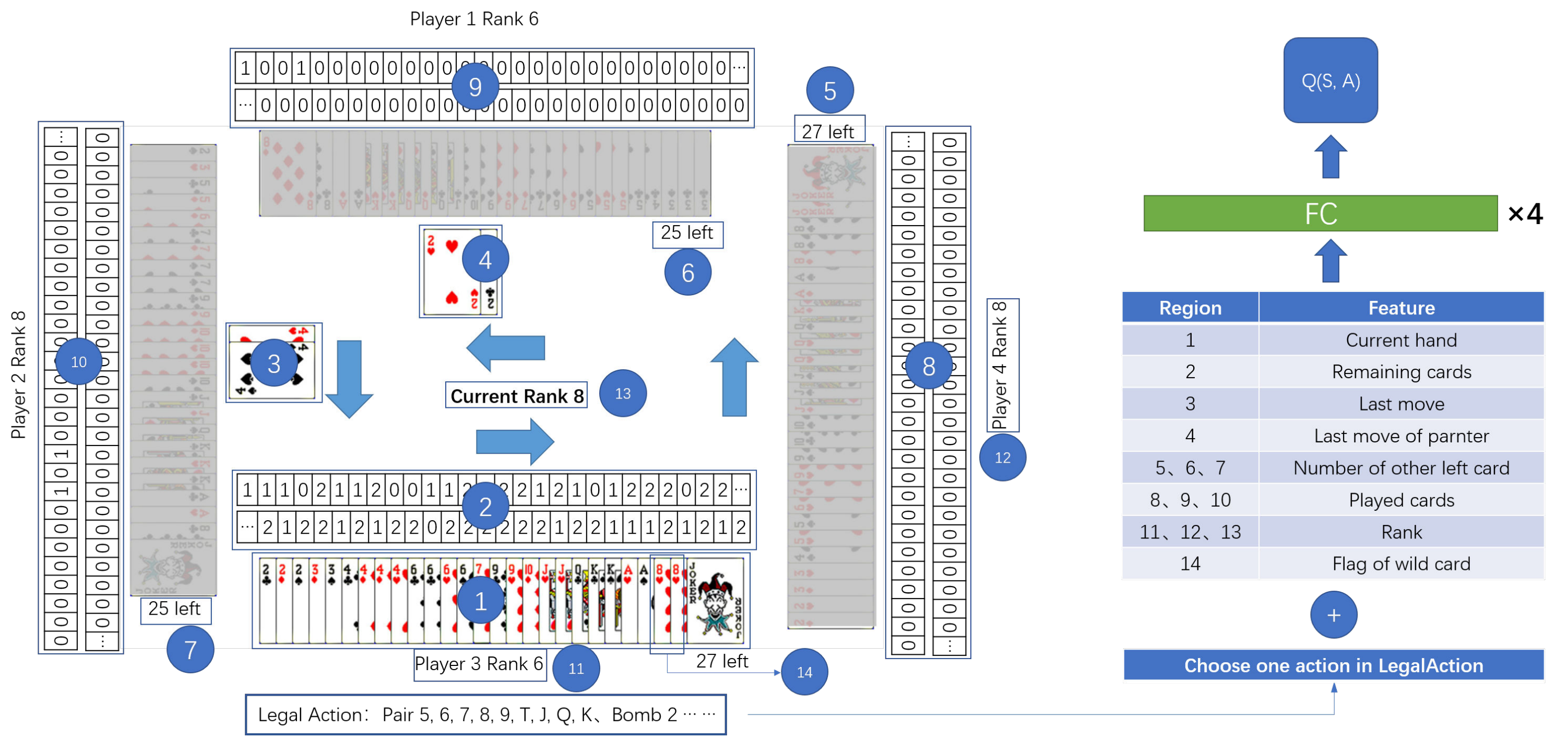}
	\caption{An example to show different regions in the GuanDan game for a better understanding of state features. The right part indicates the architecture of the network, which takes state features and legal actions as input and outputs the Q-value of a state-action pair. The features corresponding to the number area on the left are explained in the right table.}
	\label{fig4}
\end{figure*}

\subsection{Deep Monte Carlo}
As a conventional technique in the domain of reinforcement learning, the Monte Carlo (MC) method acquires knowledge of value functions and optimal policies through the analysis of experiences, involving data derived from states, actions, and rewards sampled during interactions with the environment \cite{sutton2018reinforcement}. 
Capitalizing on the advancements in neural network technologies, the MC method frequently employs deep neural networks for effective function approximation, leveraging their outstanding representational capabilities. 
This adaptation has given rise to the contemporary research methodology known as Deep Monte Carlo (DMC).

DMC is tailored for episodic tasks, characterized by experiences that can be discretely segmented into episodes with finite lengths.
The methodology updates the estimation of values and policies exclusively upon the termination of an episode. 
Following the completion of each episode, the observed returns are adopted for policy evaluation, facilitating policy enhancement at states visited during the episode. 

For the GuanDan game, characterized as a typical episodic task, the suitability of the DMC is evident, 
In the practical implementation of MC methods, an epsilon-greedy policy is commonly employed to strike a balance between exploration and exploitation during sample generation.
Also, DMC is well-matched for adoption within a distributed training framework, allowing for parallel actor execution, which can enhance the efficient collection of data samples and mitigate issues related to variance.
Moreover, the average return in DMC is often achieved by the discounted cumulative reward.
Notably, the approximation of the target Q-value in DMC is conducted without bias, diverging from Q-learning which relies on bootstrapping methodologies, making DMC avoid the overestimating issue suffered by deep Q-learning.
This method can also leverage action features to help learning while traditional policy gradient algorithms usually fail to utilize this knowledge.
Given all these advantages, we incorporate the DMC method in the development of our GuanDan AI system.

\subsection{Proximal Policy Optimization}
\label{cha:ppo}
PPO, an eminent policy gradient actor-critic algorithm, has evolved from the Trust Region Policy Optimization (TRPO) framework \cite{schulman2015trust}.
Its inception stems from addressing a critical challenge observed in the Vanilla Policy Gradient (VPG) method \cite{sutton2000policy}, namely the inherent instability in the optimization process which may inadvertently result in the collapse of the current policy's performance. 
Diverging from TRPO, which involves intricate matrix calculations, PPO circumvents this issue by employing a first-order method. 
Specifically, PPO introduces a surrogate optimization function to replace the optimization of $J(\pi)$ in VPG Optimization:
\begin{equation}\label{ppo}
\begin{split}
L^{clipped}=&\mathbb{E}_{s\sim\rho_{old},a\sim\pi_{old}}[\min(\frac{\pi_{\theta}(a|s)}{\pi_{\theta_{old}}(a|s)}\hat{A}(s,a),\\
&clip(\frac{\pi_{\theta}(a|s)}{\pi_{\theta_{old}}(a|s)},1-\epsilon,1+\epsilon))\hat{A}(s,a)],
\end{split}
\end{equation}
\begin{equation}\label{eqclip}
clip(x, x_{min}, x_{max}) = \left\{
\begin{array}{ll}
x_{max},   & x > x_{max}\\
x,      & x_{mix}\leq x\leq x_{max}\\
x_{min},   & x < x_{min} \\
\end{array}
\right. ,
\end{equation}
where $\rho_{old}$ represents the state distribution on policy $\pi_{old}$ and $\hat{A}$ is the estimated advantage.
In this context, PPO employs the clip method to delimit the divergence between the updated and previous policy.
This approach serves as an effective remedy for the problem of the policy deteriorating during training, which results from an inappropriate update step size.
The clip function, defined in Equation \ref{eqclip}, plays a pivotal role in this constraint mechanism. 
Additionally, the approximate advantage function, denoted as $A(s, a)$, is always estimated through the application of Generalized Advantage Estimation (GAE) \cite{schulman2015high}.

PPO adopts the deep neural network to approximate both the policy and state value function.
During the update of network parameters to maximize the above surrogate optimization target, an additional regularization term is introduced by incorporating the information entropy of the current policy, denoted as $H_{p}(\pi)$, which is aimed to encourage exploration.
In parallel, the parameters of the value network undergo updates through regression to the mean cumulative rewards, denoted as $R$. 
Overall, the policy loss of PPO is:
\begin{equation}\label{eqpolicy}
L_{p}=-L^{clipped}-c_{e}H_{p}(\pi),
\end{equation}
where $c_{e}$ is the entropy coefficient. And the value function loss is:
\begin{equation}\label{eqvalue}
L_{v}=\mathbb{E}_{s\sim \rho_{old}}[(V_{\phi}(s)-R)^2].
\end{equation}
The total loss of PPO is defined as follows:
\begin{equation}\label{eqtotal}
L(\theta,\phi)=L_{p} + c_{v}L_{v}.
\end{equation}

In general, PPO allows for efficient parallelization of training and typically exhibits improved sample efficiency compared to other policy gradient methods.
It also has demonstrated robust performance across a variety of environments and tasks.
These advantages make this algorithm popular in reinforcement learning problems.

\section{Method} \label{method}
Following the introduction of the basic rules of the GuanDan game, this section provides an exposition of methods adopted in our AI system design.
To be specific, we first need to obtain a GuanDan AI program trained using Deep Monte Carlo (DMC) method.
Here we introduce the state and action features design as well as the distributed training framework and the implementation of DMC method.
Subsequently, we describe how we employ the Proximal Policy Optimization (PPO) algorithm in this game, overcoming the challenges stemming from extensive action space.
Notably, these technologies are all tailored to deal with the card-playing phase of the GuanDan game, we will also present how our AI system handles the tribute phase.

\begin{figure*}[t]
	\centering
	\includegraphics[width=0.95\textwidth]{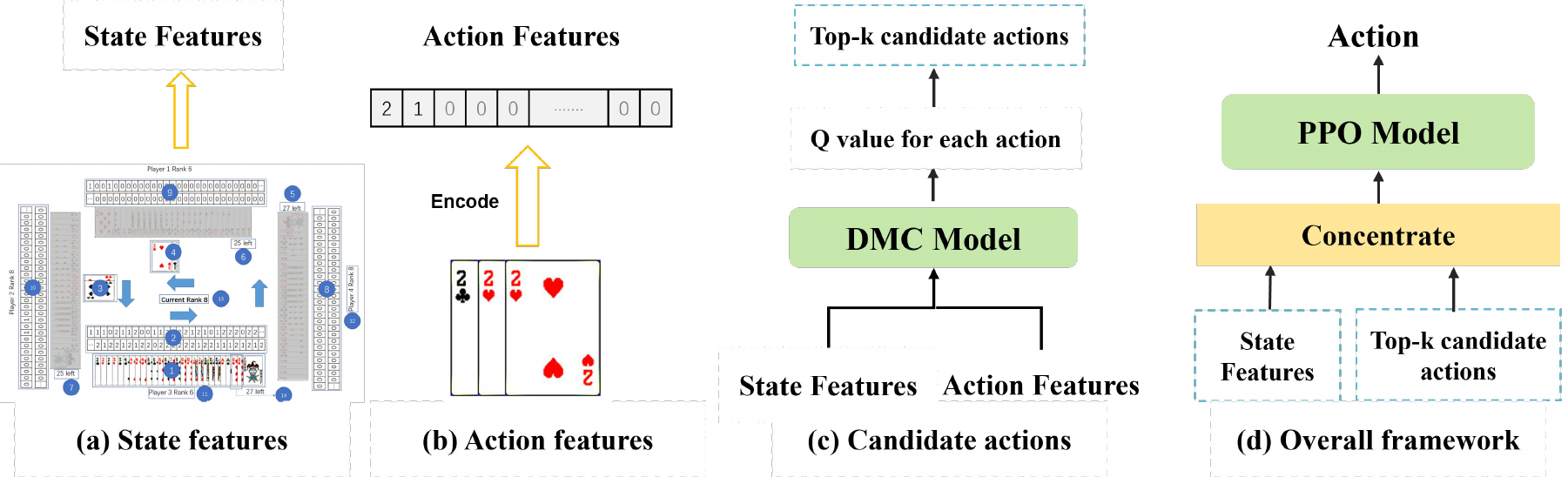}
	\caption{An overview of the framework that we enhance our GuanDan AI program using the PPO algorithm and how different features are obtained.
    The DMC model takes the state and action features as input and outputs the Q value for each value.
    We select the top-k actions with highest Q values as the candidate actions and 
	concentrate them with state features to serve as the input of PPO model.
    In this way, our method can greatly constrain the extensive action space of GuanDan game and make the application of PPO algorithm feasible.
    }
	\label{ppo_framework}
\end{figure*}

\subsection{State and Action Features Design}
A pivot facet of our model architecture revolves around the meticulous incorporation of pertinent information and candidate actions, resulting in the generation of state-action values.
To achieve this, we utilize a 54-dimensional vector encoding for each card combination, which can represent the entire deck of 54 cards in poker.
In our encoding scheme, each element in the vector has three possible values, \emph{i.e.} $\{ 0, 1, 2 \}$, denoting the number of cards belonging to the respective suits and points.
Figure~\ref{fig3} demonstrates an example of our encoding scheme.
Through this approach, we successfully embed the card information within the feature, striking a balance between informativeness and manageable dimensionality.

In implementation, as each action is a combination of cards, the action features are succinctly portrayed through the 54-dimensional vector introduced above.
Regarding the state feature of our model, it comprises a 513-dimensional vector, delineating various aspects relative to one player's perspective, with each dimension holding a specific meaning:
\begin{itemize}
    \item $\left[ 0 - 53 \right]$: The cards currently held by the player.
    \item $\left[ 54 - 107 \right]$: The remaining cards, composed of all cards excluding the player's current hand and those already played.
    \item $\left[ 108 - 161 \right]$: The last move of players and the cards to be played must be able to cover this combination of cards. 
    If the player is leading the trick, these dimensions are set to zero.
    \item $\left[ 162 - 215 \right]$: The last move made by the partner.
    If the partner's move is ``pass'', this vector is set to zero and if the partner has exhausted his hand cards, these dimensions are set to -1.
    \item $\left[ 216 - 299 \right]$: The number of remaining cards for the other three players in the order of their card play.
    \item $\left[ 300 - 461 \right]$: The played cards of the other three players, recorded in the order of playing cards.
     \item $\left[ 462 - 501 \right]$: The level of both our team and the opponent team. (13 dimensions for each: ours, theirs, and the current level)
     \item $\left[ 501 - 513  \right]$: Flags of wild cards in the player's hand and their suitability for forming various card combinations, excluding Singles and Joker Bombs.
\end{itemize}

\subsection{Distributed Training Framework}
In this subsection, we present the intricacies of our distributed training framework, which is bifurcated into two integral components: the actor and the learner.
To be mentioned, our GuanDan models are all trained using this framework, which facilitates parallelism among multiple actor processes so that the training process of our AI system is efficient.

The actor plays a pivotal role in conducting game simulations and gathering data samples.
Within each iteration, every actor is furnished with the latest model parameters from the learner.
Subsequently, the environment initiates a fresh episode.
Notably, each actor maintains four agents to interact with the game core.
In other words, in the adopted self-play procedure, each episode will produce four trajectories.
At each time step, each agent $i$ receives the preceding moves of the other three players $a_{-i}$ and updates the state feature $s_i$.
Predicted on the current state, the set of legal actions $A$ can also be computed.
With the input, the model will give corresponding output and select an action $a$, which is sent to the game core to propel the game process. 
This sequence of operations is iterated by each player until the end of an episode.

\begin{table*}[]
\resizebox{\linewidth}{!}
    {
\begin{tabular}{|cc|c|c|c|}
\hline
\multicolumn{2}{|c|}{}                                           & Deep Monte Carlo                        & Proximal Policy Optimization & Ours(DMC+PPO)   \\ \hline
\multicolumn{1}{|c|}{\multirow{2}{*}{Actor}}   & Sample ways     & $\epsilon$-greedy                                & Sample by probability        & Combined        \\ \cline{2-5} 
\multicolumn{1}{|c|}{}                         & Time Complexity & $O(n)$                                    & $O(n)$                      & $O(n)+O(k) \approx O(n)$  \\ \hline
\multicolumn{1}{|c|}{\multirow{2}{*}{Learner}} & Update ways     & $[Q(s, a) - r]^2$ & $\min(\frac{\pi_{\theta}(a|s)}{\pi_{\theta_{old}}(a|s)}\hat{A}(s,a), clip(\frac{\pi_{\theta}(a|s)}{\pi_{\theta_{old}}(a|s)},1-\epsilon,1+\epsilon))\hat{A}(s,a)$      & Combined        \\ \cline{2-5} 
\multicolumn{1}{|c|}{}                         & Time Complexity & $O(1)$                                    & $O(n)$                   & $O(1)+O(k) \approx O(k)$ \\ \hline
\end{tabular}
}
\caption{Analysis of the methods adopted in our work. 
      As for the complexity, the notation ``O(n)'' denotes the necessity for the model to explore the legal action space, posing feasibility challenges during batch processing on the learner side.
       In our approach, denoted as ``Ours'', we augment our AI program by firstly training a model using Deep Monte Carlo (DMC) to screen and identify the top-k actions. 
       Subsequently, we employ Proximal Policy Optimization (PPO) to select actions from this reduced set of candidates, which can markedly mitigate computational costs.
       The terminology ``Combined'' is assigned to signify that our method integrates both DMC and PPO techniques so that the sample ways and update ways are the combinations.
       }
       \label{compare}
\end{table*}

Following the end of a round in a GuanDan game, we attribute a value for each sample contingent upon the outcome of the round.
To elucidate, for the victorious team, the trajectories of their samples receive values of +3, +2 and +1 when the partner of the Banker is the Follower, the Third and the Dweller, respectively.
Conversely, the samples of the losing team are assigned corresponding negative values.
To be noted, we integrate a global reward with the reward of each individual round.
That is to say, if a team ultimately wins this episode of the game, we will additionally add a value of +1 to the samples in this trajectory, otherwise, we will add -1. 
To this end, the agent is encouraged to learn to achieve victory in the GuanDan game instead of paying too much attention to the victory or defeat of just a round.
Once an episode terminates, the agent trajectory data tuple is transmitted to the learner for model training.

The learner takes the responsibility of updating the network during the training process, wherein it acquires the episode data gathered by the actors and stores it in a dedicated buffer.
Subsequently, the learner samples a batch of data from the buffer to update the network using the training algorithm.
Meanwhile, although each actor maintains four agents interacting with the environment to generate data, they all utilize the same model.
This training setup aims to cultivate expertise with the AI program, enabling it to handle diverse situations encountered in this game.
The overall training framework is outlined in the appendix.

\subsection{Deep Monte Carlo Method}
In this part, we introduce the implementation of the DMC method within our distributed training framework.
On the actor side, the model takes $s$ and each legal action $a$ as input, yielding the state-action value $Q(s_i, a)$.
The action is then selected through $\epsilon$-greedy policy.
To be specific, an action is chosen randomly from the legal action set with a probability of $\epsilon$ and there simultaneously exists a probability of $1-\epsilon$ to opt for action with the highest $Q(s_i, a)$.
Following the aforementioned methodology to assign rewards for each data tuple, resulting in the tuple $(s, a, Q(s, a), r)$, the actors transmit the data samples to the learner for model training.
For the learning process, Deep Monte Carlo proves to be an efficacious value-based algorithm, particularly well-suited for episodic and reward-sparse tasks characterized by a substantial state and action space.
In this algorithm, the neural network is updated using the optimization function as follows:
\begin{equation}\label{loss}
Loss= \frac{1}{N} \sum_{i=i}^N [Q(s,a;\theta_{l})  - r)]^2, 
\end{equation}
where $Q(s,a;\theta_{l})$ refers to the state-action value predicted by the $\theta$ parameter on the learner side.
Subsequently, every actor acquires the latest model parameters from the learner and iterates through the distributed training process.

As for the model architecture of the model trained using the DMC algorithm, the employed neural network encompasses multiple layers of Multi-Layer Perception (MLP), with the input being the concatenation of the state and action features, thereby constituting a 567-dimensional vector.
The output of the network corresponds to the Q-value of a particular state-action pair.
The architectural delineation of the network and the specific segmentation forming the state vector are visually elucidated in Figure~\ref{fig4}.

\subsection{Proximal Policy Optimization}
Utilizing Deep Monte Carlo, we can achieve a GuanDan AI with good performance.
Inspired by the achievements of policy gradient reinforcement learning algorithms in extensive gaming scenarios, we further enhance our AI program by incorporating the Proximal Policy Optimization (PPO) algorithm.
Whereas, as is shown in Figure~\ref{fig1}, the GuanDan game exhibits an expansive action space and legal action space, posing a great challenge for policy gradient reinforcement learning algorithms.
Even in another imperfect information card game, DouDizhu, which is less complex compared to GuanDan, the classical policy-based reinforcement learning algorithm A3C \cite{mnih2016asynchronous} fails to achieve a satisfactory performance \cite{you2020combinatorial}.
Moreover, the requirement for computing the probability for each action, namely $\pi(a|s)$, during the training phase of policy-based reinforcement learning algorithms introduces a formidable computational obstacle.
To obtain the probability on the learner side, the cost of batch computation is unacceptable in implementation.
To elucidate, in a data batch containing diverse samples, even most states have just a few legal actions, once there exists a state with multiple legal actions, the size of the entire data batch has to be expanded to encompass the whole action space.
These intricacies collectively pose substantial hurdles to the direct application of PPO or other policy gradient reinforcement learning algorithms to this game.

To address the challenge posed by the extensive action space of this game, we utilize an approach involving the utilization of a pre-trained model, derived through the Deep Monte Carlo methodology.
The pre-trained model is used to filter out some actions with the highest Q-values, forming a subset of candidate actions.
Subsequently, we employ the PPO algorithm to train a model tasked with selecting actions from among these identified candidates.
We demonstrate the framework of our method in Figure~\ref{ppo_framework}.
Also, we give some analysis of different methods in Table~\ref{compare} and it's noted that replacing PPO with other policy gradient methods does not affect the complexity.
From the results, the direct application of policy-based methods emerges as computationally infeasible on the learner side due to the expansive action space.
In contrast, our proposed approach involves having the model select actions from the top-k choices recommended by the pre-trained model, resulting in a substantial reduction in computational complexity. 
Moreover, models trained with Deep Monte Carlo employ an epsilon-greedy policy for action sampling on the actor side while policy-based methods sample actions based on the action probability, thus facilitating better exploration during the action sampling process.
Incorporating PPO into our AI program is also partly motivated by this advantage.

Regarding the implementation details, the PPO model shares the same state input configuration as the DMC model.
Given that the PPO algorithm requires both a policy network and a value network, we design a neural network composed of multiple layers of MLP featuring two output branches.
Incorporating the knowledge of pre-trained information into the PPO model is very helpful for its learning process.
We introduce the filtered actions into the network input as the action feature. 
Then the model outputs the logits for each action, which can be used for the action probability distribution computation, and the estimated value of current state for subsequent updates.
To clarify, when the pre-trained model filters out $k$ candidate actions, the input of PPO model becomes $(s, a_1, \cdots, a_k)$ and the output comprises $(logits(a_1), \cdots, logits(a_k), V(s))$.
To be mentioned, for states with fewer than k legal actions (\emph{e.g.}, $m$ legal actions where $m < k$), we set the dimensions of the remaining actions ($a_{k-m+1}, \cdots, a_k$) to -1, a value distinct from any action encoding.
What's more, an additional feature is introduced to identify legal actions, ensuring that the probability of illegal actions is zero, thus preventing the model from selecting unreasonable actions.
Moreover, we can also employ the distributed reinforcement learning framework in implementation, enabling a highly efficient training process.
Although the PPO algorithm is chosen in our work, other policy gradient reinforcement learning methods can also be utilized.
The idea of leveraging the pre-trained model as a ``teacher'' to guide policy-based learning is transferable to analogous imperfect information games.

\subsection{Handling the Tribute Phase}
In the intricate gameplay of GuanDan, particularly in the Tribute Phase that ensues from the second round onward of each episode, a unique set of rules dictates the exchange between the Dweller and the Banker. 
As is introduced in Section~\ref{rules}, the logic for this phase is different from conventional card-playing logic, presenting a formidable challenge in training a model which is capable of handling both the tribute and card-playing phases concurrently.
What's worse, it can be hard to discern the impact of decision-making during the Tribute Phase on the ultimate outcome of a GuanDan game, rendering the training of a model for this specific phase a daunting task.
To navigate this complexity, we employ heuristic rules meticulously designed to steer decision-making during this stage.
To be mentioned, as heuristic rules are employed during the tribute phase, data samples generated in this stage are not retained in the implementation.
Here we describe the rules that we adopt to deal with the tribute phase below.

The payment of Tribute by the Dweller is defined by the game rules, leaving us to just consider the strategic selection of the card for the tribute return.
This decision-making process revolves around two critical considerations.
First, the player should seek to preserve the strength of their hand and the integrity of certain special card combinations.
Simultaneously, the player had better minimize any inadvertent augmentation of the strength or integrity of the opponent's hand cards. 
In alignment with these principles, the player ought to strategically combine their cards, assigning priority to the creation of Bombs and Straight Flush combinations.
After then, if there exists a card with a point lower than 10 that is capable of only forming a Single card type, it is prioritized for tribute return.
In the absence of such a card, the player needs to dismantle Triples or Pairs with points lower than 10.
In the event that these combinations are also unavailable, the player has to disassemble Bombs to fulfill the Tribute obligation.

\begin{table*}[htbp]
    \centering
    \resizebox{\linewidth}{!}
    {
        \renewcommand{\arraystretch}{1.1}
        \begin{tabular}{|c|c|c|c|c|c|c|c|c|c|c|}
        \hline
            Win Rate ($\%$) & baseline8 & baseline7 & baseline6 & baseline5 & baseline4 & baseline3 & baseline2 & baseline1 & Ours- & Ours \\ \hline
baseline8 & -   & 26.72    & 1.14         & 100.00         & 0.00        & 13.10         & 0.00       & 0.00         & 0.00        & 0.00         \\ \hline
baseline7 & 73.28   & -    & 42.52         & 100.00         & 6.93        & 15.31         & 0.00       & 0.00        & 0.00         & 0.00        \\ \hline
baseline6 & 98.86   & 57.48    & -        & 97.67        & 18.40        & 47.96        & 28.19        & 15.64         & 0.00        & 0.00        \\ \hline
baseline5 & 0.00   & 0.00    & 2.33         & -         & 0.00         & 0.00         & 0.00         & 0.00         & 0.00         & 0.00         \\ \hline
baseline4 & 100.00  & 93.07   & 81.60        & 100.00        & -        & 83.15         & 36.42       & 55.71        & 17.32        & 12.55       \\ \hline
baseline3 & 86.90   & 84.69    & 52.04        & 100.00        & 16.85         & -        & 12.89       &  14.28       & 0.00        & 0.00        \\ \hline
baseline2 & 100.00  & 100.00   & 71.81        & 100.00         & 63.58        & 88.19        & -       & 54.44        & 25.33       & 17.39       \\ \hline
baseline1 & 100.00  & 100.00   & 86.36         & 100.00        & 54.29        & 86.78        & 45.56       & -        & 12.77       & 9.82       \\ \hline
Ours-      & 100.00  & 100.00    & 100.00        & 100.00        & 82.68       & 100.00        & 74.67       & 87.23       & -       & 46.55       \\ \hline
Ours       & \textbf{100.00}   & \textbf{100.00}   & \textbf{100.00}       & \textbf{100.00}        & \textbf{87.45}       & \textbf{100.00}        & \textbf{82.61}       & \textbf{90.12}       & \textbf{53.45}       &  -       \\ \hline
        \end{tabular}
    }
    \caption{The average performance of the compared algorithms by playing 1000 episodes of GuanDan. The win rate of each row is achieved by test between the bot in the first column against other algorithms. ``Our-'' represents the abrasive model that removes the flag for wild cards. The results of ``Our'' and ``Our-'' are achieved after training for 30 days.}
    \label{tab1}
\end{table*}

\section{Experiments}
In this section, we undertake experimental evaluations to assess the performance of our AI system against state-of-the-art rule-based methods using the GuanDan benchmark.
Moreover, to provide a more intuitive illustration of the efficacy of our AI, we give some case studies and conduct comparisons with human players.
Here we first demonstrate the good performance of the model trained using the Deep Monte Carlo method.
Then the improvement of employing Proximal Policy Optimization to enhance our AI program is presented.
The training of our AI system is executed on a server featuring 4 Intel(R) Xeon(R) Gold 6252 CPUs @ 2.10GHz and a GeForce RTX 3070 GPU, operating within the Ubuntu 16.04 system.

\begin{figure}[t]
	\centering
	\includegraphics[width=0.9\columnwidth]{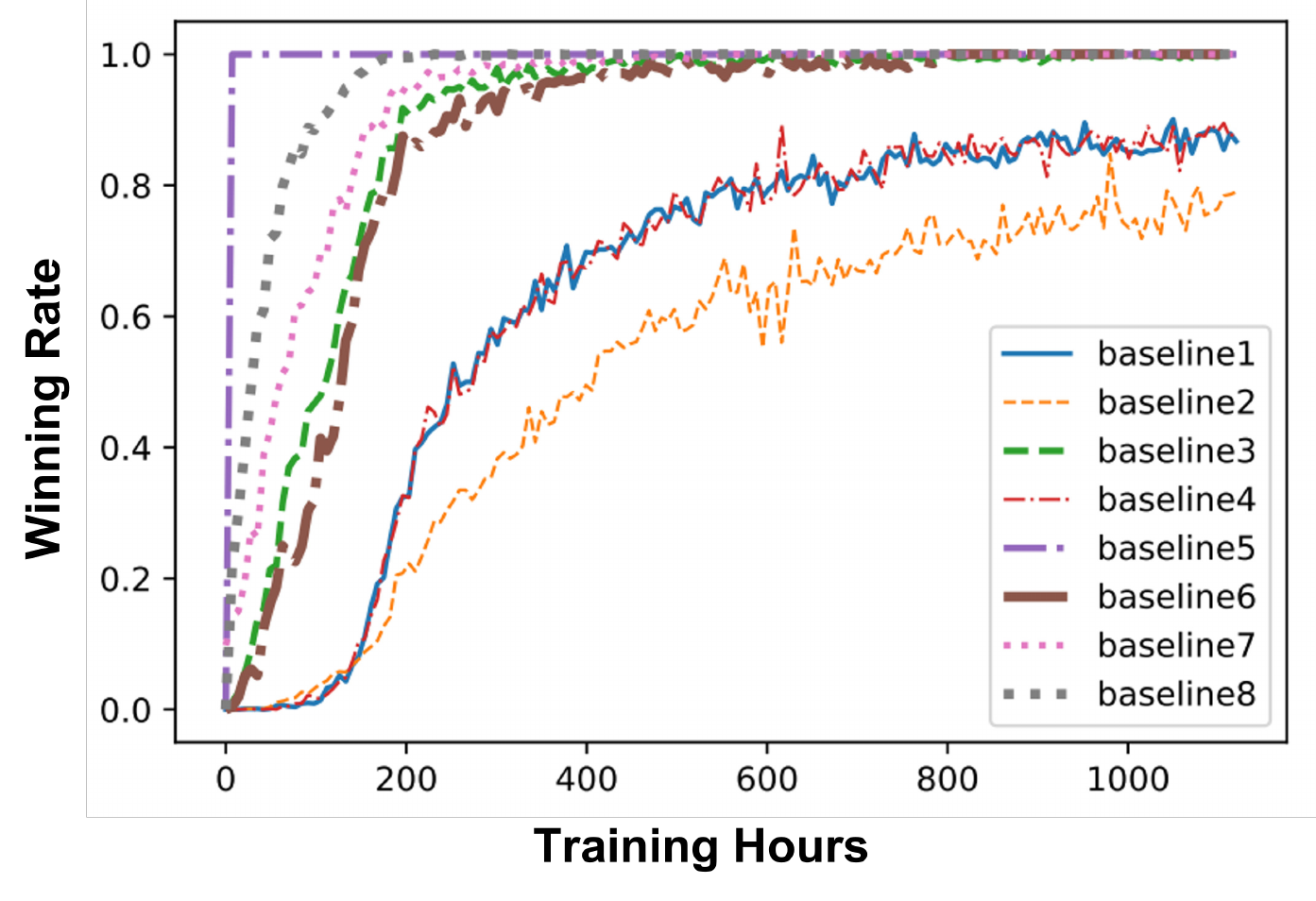}
	\caption{Winning rate for our algorithm against 8 rule-based methods. The horizontal axis represents training time of models and the vertical axis indicates the win rate of our model against rule-based bots. Evaluation against each baseline is executed with 1000 games every 24 hours. (Best viewed in color.)
	}
	\label{fig5}
\end{figure}

\subsection{Experiment Settings}
To assess the effectiveness of our AI system, we launch tournaments pitting our model against various baseline algorithms. 
Concretely, two agents belonging to one team in the GuanDan game employ our model, while the opposing team utilizes a baseline algorithm. 
For models trained using DMC, checkpoints are saved at 24-hour intervals and evaluations are conducted by executing the test for 1000 games against 8 rule-based agents.
Given the multi-round nature of a GuanDan game, playing 1000 games is sufficient to objectively reveal the performance of an AI program.
Notably, the baseline rule-based bots used for comparison are the top 8 agents in the first Chinese Artificial Intelligence for GuanDan Competition so their strengths can be guaranteed. 
Their implementations can be accessed at this website \footnote{http://gameai.njupt.edu.cn/gameaicompetition/guandan\_machi\\ne\_code/index.html}.
As for the models enhanced by PPO, owing to their training built upon the pre-trained model, the learning process is notably expedited compared to DMC models.
Consequently, checkpoints are saved every hour and we execute corresponding evaluations.

The implementation is based on our designed distributed reinforcement learning framework and the codes are available at https://github.com/submit-paper/Danzero\_plus.
An overview of hyperparameters for each method is presented in the appendix for reference.

\subsection{Performance of DMC Model against 8 Rule-based Bots and Each Other}
In this part, we present the average win rate of our model trained using the DMC method against baseline algorithms in Figure~\ref{fig5}, and the indices of the baselines represent their ranking in the competition.
The results illustrate that our AI system consistently outperforms the rule-based agents.
Notably, our AI has absolute superiority over the rule-based bots except for baseline 1, baseline 2 and baseline 4 after sufficient training.
Furthermore, it's to be noted that the initial distribution of hand cards imparts a high degree of variance to the game outcomes.
Despite this variability, DanZero attains a conspicuous advantage by achieving an 80\% win rate, signifying its superior performance compared to the other three baseline algorithms.

An intriguing observation emerges from our experiments: contrary to expectations,  baseline 2 achieves the most favorable outcome against our model and baseline 3 appears to underperform compared to baseline 4 and baseline 6.
To delve into this phenomenon, comprehensive evaluations are conducted among all AI programs and the results are demonstrated in Table~\ref{tab1}.
Remarkably, the performance of different baselines does not precisely align with their ranked positions after thorough evaluation, contributing to the observed anomalies.
To be specific, the overall performance of baseline 1, baseline 2, and baseline 4 appears equivalent.
Whereas, baseline 2 notably outperforms the other two when pitted against DanZero.
We assume that the intricate dynamics among different policies underlie this phenomenon.
In fact, a seemingly weaker bot can triumph over a stronger adversary if its policy is adept at exploiting the opponent's vulnerabilities, which is common in AI programs employing heuristic rules. 
Nevertheless, our AI system exhibits robust adaptability across agents with diverse styles, asserting its effectiveness and substantial superiority.

Besides, we assess the efficacy of the indicator for ``wild cards'' in the state feature.
An abrasive experiment is performed by eliminating these dimensions in the state features, and the results are presented in Table~\ref{tab1} denoted as ``Our-''.
The outcomes reveal that the removal of this information adversely impacts the performance of our model, indicating the inclusion of this feature enhances the model's ability to comprehend the utilization of ``wild cards''.

\begin{figure}[t]
	\centering
	\includegraphics[width=0.91\columnwidth]{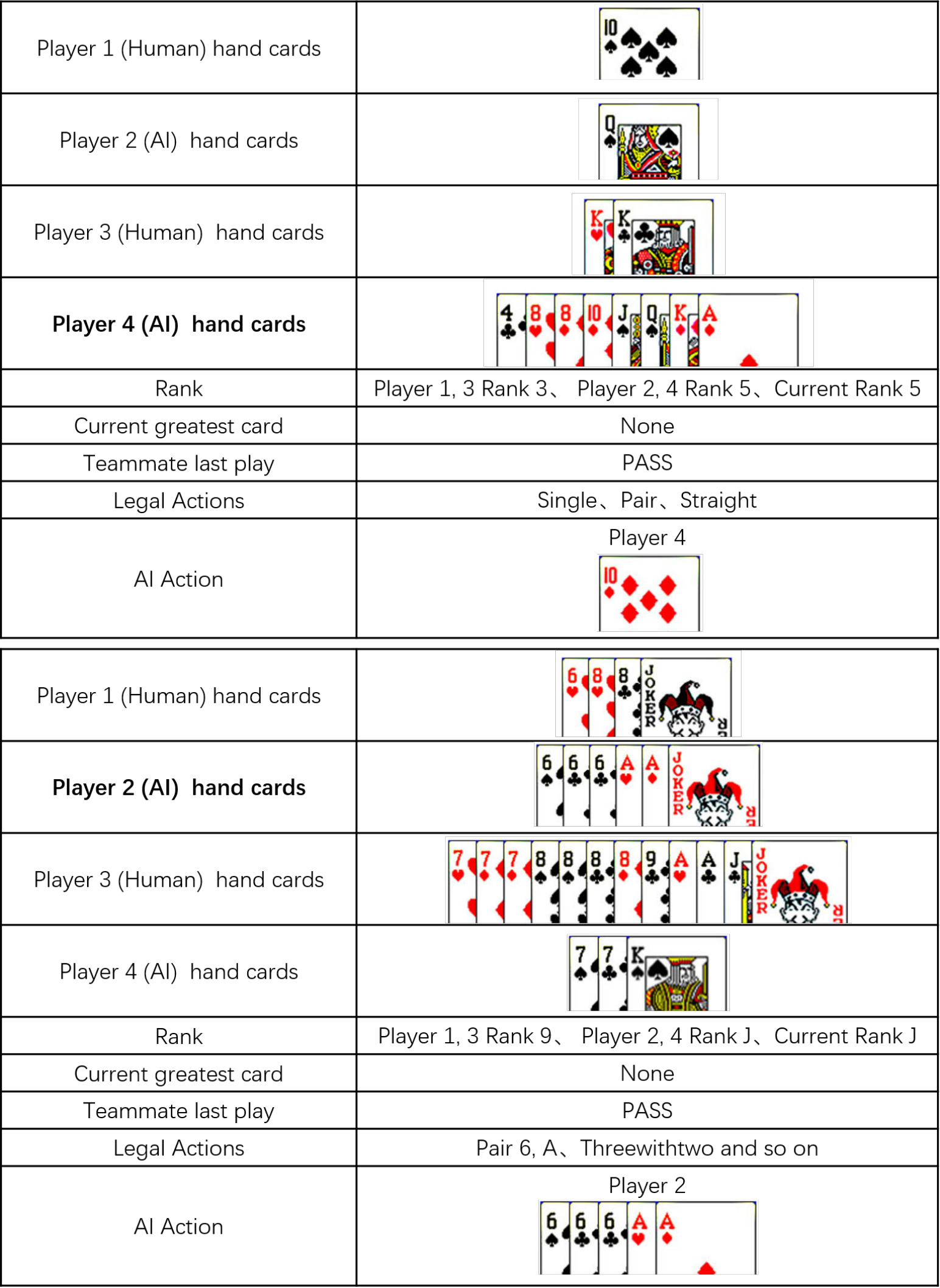}
	\caption{Case study to show skills of DanZero. The player1 and player3 are human players while player2 and player4 are controlled by our DanZero. The bold font indicates that it is the player's turn to play cards. The row of Rand indicates the levels of both teams and the current round. The current greatest card means the last move in this trick and the cards that the current player is going to play must cover it. The table also reports the last move of the partner and legal action set. The row of AI Action is the actual action that the bot decides to take.}
	\label{fig6}
\end{figure}

\subsection{Human Evaluation of DMC Model}
In addition to comparisons with strong rule-based adversaries, we extend our evaluation to assess the real performance of DanZero against human players.
Here, ten proficient graduate students, adept in flexible reasoning, are invited to participate. 
While not professional Guandan players, these individuals possess a certain level of gameplay skill and a good understanding of the game's rules. 
The competition involves pairs of human players collaborating as teammates to engage in twenty rounds against two AI opponents.
During the evaluation, our AI program achieved victory in 71 rounds out of 100 games. 
Through these human-AI gameplays, some instances show that our AI program demonstrates sound decision-making capabilities and we provide some case studies to further elucidate the prowess of our AI system in Figure~\ref{fig6}. 

In the first case depicted in Figure~\ref{fig6}, it is player4's turn to make a card play.
The permissible action set comprises Single, Pair, and Straight, while player1 and player2 hold only one card each, while player3 retains two cards. 
The state feature incorporates information on the remaining cards, enabling player4 to discern the existence of one 10, one Q, and two K in the remaining deck.
Faced with this context, playing a Single 4, Pair 8, and Straight is fraught with a high likelihood of failure. 
Consequently, the agent opts to strategically dismantle the potential Straight combination and instead plays a Single 10 to enhance the chances of its team's success.

\begin{figure}[t]
	\centering
	\includegraphics[width=0.9\columnwidth]{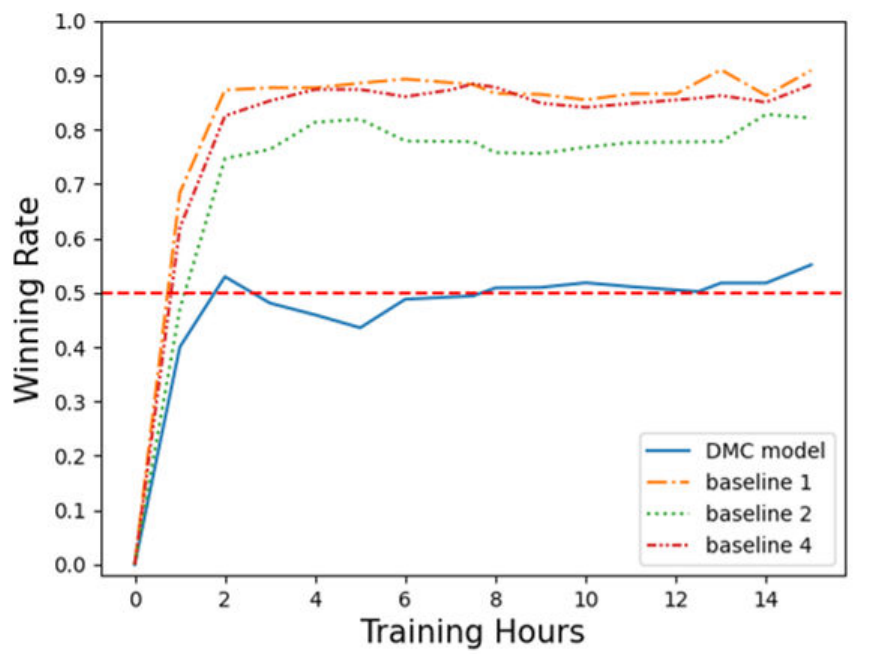}
	\caption{Winning rate for our algorithm against the DMC model and three baseline algorithms. The horizontal axis represents training time of models and the vertical axis indicates the win rate of our model. Evaluation against each baseline is executed with 1000 games.(Best viewed in color.)
	}
	\label{result2}
\end{figure}

In the second scenario, player2 needs to make a card play decision, and the legal action set encompasses Pair, Full House, and analogous combinations.
Despite the teammate possessing a mere three cards, player2 does not choose self-sacrifice for their partner's victory.
Concurrently, player3 maintains a substantial card count, and the model's awareness of the remaining 7 and 8 suggests the likelihood of bomb combinations in player3's hand.
Consequently, player2 refrains from playing the Red Joker, recognizing the strategic value in preserving it for opportune use.
In this context, player2 opts to play a Full House, retaining the Red Joker in hand. 
Anticipating the presence of remaining Single cards, this decision positions player2 favorably for subsequent plays, thus raising the likelihood of clinching victory.

From the aforementioned case analyses, it is evident that our AI program exhibits a learned capacity to discern instances for collaborative gameplay or a more individualistic style aimed at victory.
This nuanced decision-making ability underscores the high-level skill of our model in the GuanDan game. 

\subsection{Evaluation of Enhanced AI program}
In our pursuit of advancing the capabilities of our AI system, we employ the final model derived from DMC as the pre-trained model.
To get an intuitive assessment of our approach, we pragmatically set the number of candidate actions to be 2 for experiments.
Analysis about experimental outcomes of the DMC model reveals its superior performance, absolutely surpassing rule-based methodologies except baseline 1, baseline 2 and baseline 4.
For convenience, we evaluate the performance of the enhanced AI program against these three baseline algorithms and the DMC model to assess the improvement.
Despite the apparent simplification achieved by restricting the number of candidate actions in the GuanDan game to 2, the complexity of this game renders merely knowing two optimal actions insufficient. 
Notably, we evaluate a model that randomly chooses one action from the two candidates. 
This model struggles to achieve victory even against the baseline 7 opponent, achieving a winning rate of less than 10\%.

The performance of our augmented AI program is shown in Figure~\ref{result2}.
Due to the constraint of action space in our method, an expeditious acquisition of skills is evident, with the model achieving satisfactory performance in less than a day.
Notably, despite the commendable strength exhibited by the DMC model, applying the PPO algorithm yields further enhancements to our AI system.
This assertion is substantiated by the results presented in Table~\ref{tab_ppo}, wherein our enhanced AI model consistently demonstrates improved performance when compared to the DMC model.

We additionally undertake experiments to investigate the impact of the number of candidate actions. 
It is noteworthy to emphasize that in the ablation study, we maintain consistent hyperparameters.
The outcomes are presented in Table~\ref{tab_ppo}.
This method requires getting candidate actions through the DMC model, thereby escalating computational demands.
Moreover, the PPO algorithm entails more hyperparameters in comparison to DMC, such as weights for different loss functions and parameters within the Generalized Advantage Estimation (GAE) to estimate advantage values.
Consequently, the tuning process for hyperparameters demands heightened precision.
It can be observed that directly applying the hyperparameters from the PPO model trained with two candidate actions may negatively affect the performance.
Whereas, the PPO models still achieve better performance compared to the baseline bots, proving the effectiveness of our method.

\begin{table}[]
\begin{tabular}{|c|c|c|c|c|}
\hline
              & baseline1 & baseline2 & baseline4 & DMC model \\ \hline
DMC model     & 90.12     & 82.61     & 87.45     & --        \\ \hline
PPO model (2)  & \textbf{92.70}     & \textbf{86.22}     & \textbf{90.83}     & \textbf{55.13}     \\ \hline
PPO model (3) & 72.17     & 63.33     & 84.25     & 42.30     \\ \hline
PPO model (5) & 70.28     & 54.60     & 78.06     & 38.45    \\ \hline
\end{tabular}
\caption{Winning rate for PPO models under different numbers of candidate actions against the DMC model and three baseline algorithms for 1000 games. The win rate of each row is achieved by test between the bot in the first column against other algorithms. The notation of ``PPO model (xx)'' represents the PPO model trained with xx candidate actions.
	}
\label{tab_ppo}
\end{table}

\subsection{Comparison of Enhanced Model and DMC Model}
\begin{figure}[t]
	\centering
	\includegraphics[width=0.91\columnwidth]{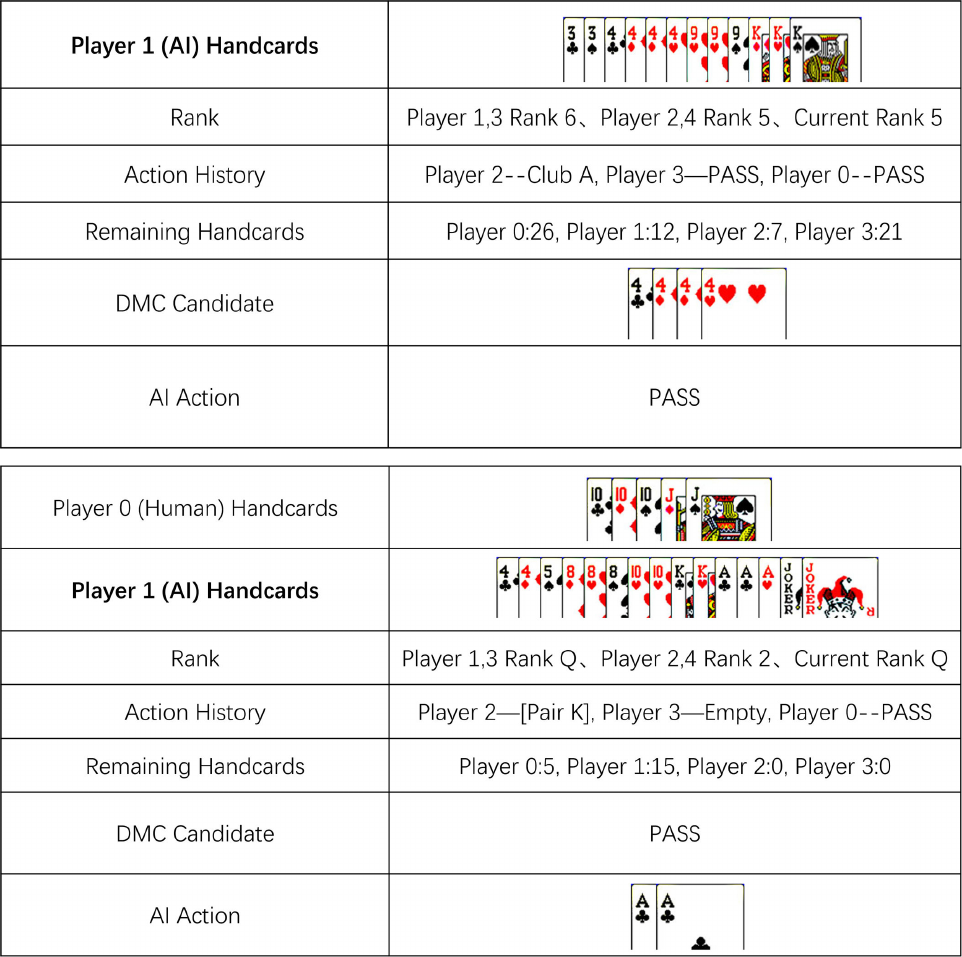}
	\caption{Case study that illuminates that the PPO model does not invariably opt for the first action suggested by the DMC model. The player0 and player2 are human players and player1 and player3 are controlled by the PPO model. The bold font signifies the active player that has to play cards. The row of ``Rank'' indicates the levels of both teams and the current round. 
   The ``Action History'' denotes the preceding move of each player in the current trick and ``Remaining Handcards'' reflects the number of cards left in each player when the current player has to make a decision.
   ``DMC Candidate'' means the top-recommended action proposed by the DMC model, whereas ``AI Action'' is actual actions undertaken by the PPO model.}
	\label{case_ppo}
\end{figure}

Given the reliance of our augmented AI program on the pre-trained DMC model, a conceivable expectation might be that the augmented model predominantly selects the most optimal candidate action advocated by the pre-trained model.
In an effort to provide a clearer demonstration of the substantive impact of the PPO algorithm on refining the model's skills within the context of the GuanDan game, we give some case studies wherein the PPO model diverges in its choices compared to the DMC model
The instances are shown in Figure~\ref{case_ppo}.

In the first scenario, it's player1's turn to make a card play.
The DMC model suggests playing Bomb 4 in this situation while PPO model opts to PASS because all four players still possess quite a few cards in hand.
Thinking it unnecessary to play the Bomb in this instance, the model elects to retain the Bomb for potential future utilization.
In the second case, the player3 has successfully emptied his hand cards and player2 also depletes his cards after playing Pair K.
The player0 chooses to PASS and the DMC model also recommends PASS. 
This choice becomes pivotal as it positions player0 to lead the ensuing trick, where he can play Full House and beat player1.
In contrast, the PPO model can better discern player0's current cards and it decisively plays the Pair A to surpass the move made by player2.

These examples indicate that the PPO algorithm illustrates the efficacy of the PPO algorithm in enhancing the model's adaptability to diverse scenarios based on available information. 
They also demonstrate that the PPO model surpasses a simplistic adherence to the top-recommended action proposed by the DMC model, exhibiting a capacity for more reasonable decision-making in response to varying contextual factors.

\section{Conclusion}
In this paper, we put forward an AI system for the GuanDan game, an imperfect information game with some significant challenges, such as the large state and action space and an indeterminate player count. 
To address these challenges, we leverage Deep Monte-Carlo (DMC) Methods as the foundational algorithm, characterizing the state information and utilizing a distributed self-play paradigm, resulting in a strong reinforcement learning agent called DanZero.
Through evaluation with state-of-the-art rule-based bots, our AI program shows outstanding performance.
To further enhance our AI program, we propose a method to apply Proximal Policy Optimization (PPO) technology, wherein the pre-trained DMC model is harnessed to constrain the extensive action space.
Our empirical results substantiate the efficacy of this approach in augmenting the performance of our GuanDan AI. 
We aspire for our contributions to serve as a benchmark, paving the way for subsequent research endeavors in the realm of the GuanDan game.

\bibliographystyle{IEEEtran}
\bibliography{refer.bib}

\newpage
\section{Appendix}

\subsection{Rules in Guandan}
In a GuanDan game, there are two decks of standard pokers used, including Jokers, and four players sitting around a square table, each of whom is dealt with a hand comprising 27 cards. 
An intriguing facet of this gameplay is the formation of two opposing camps, with players seated opposite each other being affiliated with the same faction. 
Further complexity is introduced through the notions of ``leveling up'', ``level cards'', and ``wild cards''.
To be specific, both camps of this game are on their own level which starts from 2 to A. 
The initial round of a GuanDan game invariably commences at Level 2, with subsequent round levels contingent upon the triumph of the preceding round's victorious camp.
Cards of the same rank as ongoing round's level are called ``level cards'' and they hold a rank just below Jokers when being played individually.
Moreover, these cards maintain their natural rank order when contributing to the composition of other combinations.
Particularly noteworthy are Heart cards of the level rank, which are defined as ``wild cards.''
They can substitute for any required cards in combination formation, excluding Jokers.
The overarching objective in GuanDan is for a camp to be the first to achieve a level surpassing `A', thereby clinching victory.
To this end, one GuanDan game usually encompasses multiple rounds, each marked by strategic interplay and nuanced level dynamics.

In a GuanDan game, players play cards in counterclockwise order and the player leading the first trick possesses the flexibility to play any card type from their hand.
Subsequent players are confronted with the decision to play cards of the same type or bombs, which surpass the previous player's cards, or opt for a pass. 
The continuity of a trick persists until three successive players choose to pass, at which point the last player who contributed cards will lead in the ensuing trick.
Such procedures endure until either three players have no cards left or players from the same camp empty their cards, culminating in the end of the round.
The player achieving the first card depletion is called ``Banker'', while other players are designated the ``Follower'', the ``Third'', and the ``Dweller'' based on the sequential order of card depletion.
Only the Banker's team can promote the level and the promoted number ranges from three to one, determined by the specific order of card depletion for the Banker's partner.
If the winning team empties the hand cards in the first and second positions, they will be able to promote three levels and their opponents are called the Double-Dweller.
In addition, if a player has emptied his card and other players all choose to pass in this trick, his partner will lead the subsequent trick.

From the second round onward, prior to the beginning of the initial trick, the Dweller of the preceding round is obliged to pay a Tribute to the Banker by relinquishing his biggest single card, excluding the wild card.
In reciprocation, the Banker needs to return a single card with a point not surpassing 10 to ensure that each player has the same number of cards.
Then the Dweller can lead the first trick.
In instances where there exists Double-Dweller, both players of this camp has to pay the Tribute and the winning team also needs to return the cards as discussed above.
In this process, the Banker accepts the Tribute with higher rank and his partner receives the other Tribute.
Whereas, an exception is that when the player or team obligated to pay the Tribute possesses two Red Jokers, the Tribute phase can be annulled, and the Banker takes charge of leading the first trick.

Last but not least, when the ongoing round reaches the level of Q or K and the victorious team can promote 3 or 2 levels, there exists a notable restriction: the progression cannot skip the level A.
The imposition of this constraint ensures that the level A remains an indispensable milestone in the strategic hierarchy.
Furthermore, in rounds with level A, a camp can only win when the Banker's partner is the Follower or the Third, adding more complexity to the dynamics of GuanDan.

\section{The Hyperparameters used in our AI system}
The hyperparameters that we used in our AI system is shown below.

\begin{table}[h]
    \centering
    \resizebox{\linewidth}{!}
    {
        \renewcommand{\arraystretch}{1.25}
        \begin{tabular}{|c|c|c|}
        \hline
            Hyper parameters ($\%$) & Meaning & Value \\ \hline
Epsilon & Probability of random exploration in actor & 0.01\\ \hline
Mempool size &Memory pool size in learner & 65536\\ \hline
Batch size & Batch size per training & 32768 \\ \hline
Training freq & Number of receptions between each training & 250 \\ \hline
Lamda & Range of clip Q & 0.65 \\ \hline
lr & Learning rate & 0.001 \\ \hline
Optimizer & Optimizer in training  & RMS \\ \hline
Actor num & Number of actor & 80 \\ \hline
Actor core & CPU core per actor & 2 \\ \hline
MLP layer & Number of MLP layers & 4 \\ \hline
MLP node & Number of MLP nodes per layer & 512 \\ \hline
Activation functions & Activation functions per layer & tanh \\ \hline
        \end{tabular}
    }
    \caption{The hyperparameters used in the DMC model of our AI system.}
    \label{tab_set}
\end{table}

As for the PPO enhanced model, we list the hyperparameters achieved on the setting where the number of candidate actions is 2.

\begin{table}[h]
    \resizebox{\linewidth}{!}
    {
        \renewcommand{\arraystretch}{1.25}
        \begin{tabular}{|c|c|c|}
        \hline
            Hyper parameters ($\%$) & Meaning & Value \\ \hline
Mempool size &Memory pool size in learner & 2048\\ \hline
Batch size & Batch size per training & 2048 \\ \hline
Training freq & Number of receptions between each training & 13 \\ \hline
Gamma & gamma of GAE & 0.99 \\ \hline
Lambda & lambda of GAE & 0.95 \\ \hline
lr & Learning rate & 0.0001 \\ \hline
Optimizer & Optimizer in training  & Adam \\ \hline
Actor num & Number of actor & 40 \\ \hline
Actor core & CPU core per actor & 2 \\ \hline
MLP settings & Settings of MLP network & 512,512,512,256 \\ \hline
Activation functions & Activation functions per layer & tanh \\ \hline
Action number & Number of Candidate actions for training & 2 \\ \hline
Clip ration & The value of $\epsilon$ in the clipped policy loss in PPO & 0.2 \\ \hline
Loss weights & Weights for policy loss, value loss and entropy loss & 1, 0.5, 0.05 \\ \hline
        \end{tabular}
    }
    \caption{The hyperparameters used in our AI system.}
    \label{tab_set}
\end{table}

\section{The Overall Algorithm Framework used in our AI system}
As our training of the GuanDan models is implemented through an efficient distributed training framework, here we give the outline for better understanding.

\begin{algorithm}
	\renewcommand{\algorithmicrequire}{\textbf{Input:}}
	\renewcommand{\algorithmicensure}{\textbf{Output:}}
	\caption{Process of Actor}
	\begin{algorithmic}[1]
		    \STATE Initialize environment ENV;
		    \STATE Initialize model $M$ with random parameters;
            \FOR{Episodes=1,2,3,...}
		        \STATE Initial state $s_{0}$ = ENV.reset();
		        \STATE Set $t = 0$;
		        \WHILE {not done}
    		        \FOR{Agent=1,2,3,4}
        		        \STATE Calculate the legal actions set $A$;
        		        \STATE Choose action $a^i$ according to sample policy;
        		        \STATE Collect set of trajectories by interacting with the environment:
                            \STATE $\tau^i_t=f(\tau^i_{t-1}, a^{-i})$;
        		    \ENDFOR
                    \STATE $t = t+1$;
                \ENDWHILE
                \STATE Assign a value $r$ for every sample;
                \STATE For each trajectory $(\tau_t, a_t,Q(\tau_t,a_t), r_t)$, save it to replay buffer $B$;
                \STATE Update model $M$ with period $I$;
		    \ENDFOR
	\end{algorithmic}
	\label{actor}
\end{algorithm}

\begin{algorithm}
	\renewcommand{\algorithmicrequire}{\textbf{Input:}}
	\renewcommand{\algorithmicensure}{\textbf{Output:}}
	\caption{Process of Learner}
	\begin{algorithmic}[1]
        \STATE Initialize the network parameters and replay buffer $B$;
        \FOR{Iteration=1,2,3,...}
        \STATE Sample a batch of trajectory data $D=\{(\tau, a, Q(\tau,a), r)\}$ from $B$;
    	\STATE Calculate the loss function $L(\theta)$ according to the corresponding algorithm;
    	\STATE Update state-action value network parameters $\theta$ with $L(\theta)$;
        \STATE Send network parameters to Actor;
        \ENDFOR
	\end{algorithmic}
	\label{learner}
\end{algorithm}

\end{document}